\documentclass[12pt,reqno]{amsart} 

\usepackage{mathtools,enumitem,lscape}
\usepackage{multirow}

\usepackage{graphicx,amssymb,colordvi,textcomp,latexsym,mathtools,}
\usepackage[hidelinks]{hyperref}
\usepackage{tabularx}
\usepackage{booktabs}
\usepackage{float}
\usepackage{prettyref}
\usepackage{chngcntr}
\counterwithout{equation}{subsection}
\counterwithout{equation}{section} 
\usepackage{lscape}
\usepackage{xcolor}
\newcommand{\htodo}[1]{}
\newtheorem{lemma}{Lemma}

\newtheorem{corollary}{Corollary}

\theoremstyle{plain}
\newtheorem{remark}{Remark}
\newtheorem{theorem}{Theorem}
\newtheorem{definition}{Definition}

\def\ddefloop#1{\ifx\ddefloop#1\else\ddef{#1}\expandafter\ddefloop\fi}
\def\ddef#1{\expandafter\def\csname 
	bb#1\endcsname{\ensuremath{\mathbb{#1}}}}
\ddefloop ABCDEFGHIJKLMNOPQRSTUVWXYZ\ddefloop
\def\ddef#1{\expandafter\def\csname 
	#1\endcsname{\ensuremath{\mathbf{#1}}}}
\ddefloop abcdefghijklmnopqrstuvwxyz\ddefloop
\def\ddefloop#1{\ifx\ddefloop#1\else\ddef{#1}\expandafter\ddefloop\fi}
\def\ddef#1{\expandafter\def\csname 
	b#1\endcsname{\ensuremath{\mathbb{#1}}}}
\ddefloop ABCDEFGHIJKLMNOPQRSTUVWXYZ\ddefloop
\def\ddef#1{\expandafter\def\csname 
	c#1\endcsname{\ensuremath{\mathcal{#1}}}}
\ddefloop ABCDEFGHIJKLMNOPQRSTUVWXYZ\ddefloop
\def\ddef#1{\expandafter\def\csname 
	f#1\endcsname{\ensuremath{\mathfrak{#1}}}}
\ddefloop ABCDEFGHIJKLMNOPQRSTUVWXYZabcdefghjklmnopqrstuvwxyz\ddefloop
\def\ddef#1{\expandafter\def\csname 
	h#1\endcsname{\ensuremath{\widehat{#1}}}}
\ddefloop ABCDEFGHIJKLMNOPQRSTUVWXYZ\ddefloop
\def\ddef#1{\expandafter\def\csname 
	hc#1\endcsname{\ensuremath{\widehat{\mathcal{#1}}}}}
\ddefloop ABCDEFGHIJKLMNOPQRSTUVWXYZ\ddefloop

\newcommand{\balpha}{\boldsymbol{\alpha}}

\newcommand{\bones}{\mathbf{1}}

\newcommand{\defeq}{\coloneqq}
\DeclarePairedDelimiter{\brk}{[}{]}
\DeclarePairedDelimiter{\crl}{\{}{\}}
\DeclarePairedDelimiter{\prn}{(}{)}
\DeclarePairedDelimiter{\nrm}{\|}{\|}
\DeclarePairedDelimiter{\inner}{\langle}{\rangle}

\DeclareMathOperator*{\argmin}{arg\,min} 

\newcommand{\sphere}{\cS}
\newcommand{\cl}[1]{\overline{#1}}

\usepackage{prettyref}
\newcommand{\pref}[1]{\prettyref{#1}}
\newcommand{\pfref}[1]{Proof of \prettyref{#1}}
\newcommand{\savehyperref}[2]{\texorpdfstring{\hyperref[#1]{#2}}{#2}}
\newrefformat{eq}{\savehyperref{#1}{\textup{(\ref*{#1})}}}
\newrefformat{eqn}{\savehyperref{#1}{Equation~\ref*{#1}}}
\newrefformat{lem}{\savehyperref{#1}{Lemma~\ref*{#1}}}
\newrefformat{obs}{\savehyperref{#1}{Observation~\ref*{#1}}}
\newrefformat{def}{\savehyperref{#1}{Definition~\ref*{#1}}}
\newrefformat{thm}{\savehyperref{#1}{Theorem~\ref*{#1}}}
\newrefformat{corr}{\savehyperref{#1}{Corollary~\ref*{#1}}}
\newrefformat{sec}{\savehyperref{#1}{Section~\ref*{#1}}}
\newrefformat{app}{\savehyperref{#1}{Appendix~\ref*{#1}}}
\newrefformat{ass}{\savehyperref{#1}{Assumption~\ref*{#1}}}
\newrefformat{ex}{\savehyperref{#1}{Example~\ref*{#1}}}
\newrefformat{fig}{\savehyperref{#1}{Figure~\ref*{#1}}}
\newrefformat{alg}{\savehyperref{#1}{Algorithm~\ref*{#1}}}
\newrefformat{rem}{\savehyperref{#1}{Remark~\ref*{#1}}}
\newrefformat{conj}{\savehyperref{#1}{Conjecture~\ref*{#1}}}
\newrefformat{prop}{\savehyperref{#1}{Proposition~\ref*{#1}}}
\newrefformat{table}{\savehyperref{#1}{Table~\ref*{#1}}}
\newrefformat{prob}{\savehyperref{#1}{Problem~\ref*{#1}}}
\newrefformat{claim}{\savehyperref{#1}{Claim~\ref*{#1}}}
\newrefformat{fact}{\savehyperref{#1}{Fact~\ref*{#1}}}
\newrefformat{sec}{\savehyperref{#1}{Section~\ref*{#1}}}
\newrefformat{exam}{\savehyperref{#1}{Example~\ref*{#1}}}
\newrefformat{exams}{\savehyperref{#1}{Examples~\ref*{#1}}}
\newrefformat{prob}{\savehyperref{#1}{(\ref*{#1})}}

\usepackage{setspace}

\newcommand{\RR}{\mathbb{R}}
\newcommand{\EE}{\mathbb{E}}

\newcommand{\NN}{\mathbb{N}}

	\newcommand{\ploss}{{\cL}} 
\newcommand{\Od}[1]{{{\text{O}(#1)}}}

\newcommand{\bxi}{{\boldsymbol{\xi}}}

\newcommand{\dd}{\mbox{$\;|\;$}}

\newcommand{\defoo}{~\dot{=}~}

\newcommand{\pint}{\mbox{$\mathbb{N}$}}

\newcommand{\arr}{{\rightarrow}}

\definecolor{myblue}{rgb}{0.9, 0.9, 1}

\newcommand{\WW}{W}

\newcommand{\real}{\mathbb{R}}

\newcommand{\critset}{\Sigma}
\newcommand{\tanset}{\mho}

\newcommand{\ibr}[1]{[#1]}

\title[Hidden Minima]{Hidden Minima in Two-Layer ReLU Networks}

\author{Yossi Arjevani}
\address{Yossi Arjevani, The Hebrew University, Jerusalem}
\email{yossi.arjevani@gmail.com}

\begin{document}

\begin{abstract}
We consider the optimization problem associated with training two-layer ReLU networks with \(d\) inputs under the squared loss, where the labels are generated by a target network. Recent work has identified two distinct classes of infinite families of minima: one whose training loss vanishes in the high-dimensional limit, and another whose loss remains bounded away from zero. The latter family is empirically avoided by stochastic gradient descent, hence \emph{hidden}, motivating the search for analytic criteria that distinguish hidden from non-hidden minima. A key challenge is that prior analyses have shown the Hessian spectra at hidden and non-hidden minima to coincide up to terms of order \(O(d^{-1/2})\), seemingly limiting the discriminative power of spectral methods. We therefore take a different route, studying instead certain curves along which the loss is locally minimized. Our main result shows that arcs emanating from hidden minima exhibit distinctive structural and symmetry properties, arising precisely from \(\Omega(d^{-1/2})\) eigenvalue contributions that are absent from earlier analyses.

\end{abstract}
\maketitle

\newcommand{\ba}{\boldsymbol{a}}
\newcommand{\bb}{\boldsymbol{b}}
\section{Introduction}

An outstanding question in deep learning (DL) concerns the ability of
simple gradient-based methods to successfully train neural networks despite the nonconvexity of the associated optimization problems. Indeed, nonconvex optimization landscapes may have spurious (i.e., non-global local) minima with large basins of attraction and this can cause a complete failure of these methods. Our understanding of the nature by which nonconvex problems associated with artificial neural networks differ from computationally hard ones is currently limited. In view of the complexity exhibited by contemporary networks and the absence of suitable analytic tools, much recent research has focused on two-layer ReLU networks as a realistic starting point for a theoretical study, e.g., 
\cite{brutzkus2017globally,chizat2018global,soltanolkotabi2018theoretical,mei2018mean,goldt2019generalisation,tian2020student,safran2021effects}. The two-layer networks considered were typically of the form:
\begin{align} \label{net}
	f(\x; \WW, \ba) \defeq
	\ba^\top\!\sigma\prn{W\x},\quad
	~\WW\in   
	M(k,d),~\ba \in \RR^k,
\end{align}
where $\sigma(t)\defeq\max\crl{0,t}$ is the ReLU function acting entrywise and $M(k,d)$ denotes the space of $k\times d$ matrices. To isolate the study of optimization-related obstructions on account of nonconvexity from those pertaining to the expressive power of two-layer networks, data has been often assumed to be fully realizable. This was further motivated by hardness results indicating a strict barrier inherent to the explanatory power of 
distribution-free approaches operating in complete generality 
\cite{blum1989training,brutzkus2017sgd,shamir2018distribution}. 
For the squared loss, the resulting highly nonconvex expected loss~is
\begin{align}\label{opt:problem}
	\cL(\WW, \balpha) \defeq \frac{1}{2}\EE_{\x\sim 
		\cD}\brk*{\Big(f(\x; \WW, \ba) -      f(\x; T, 
		\boldsymbol{b})\Big)^2},
\end{align}
where $\cD$ denotes a probability distribution over the input space, $\WW\in M(k,d)$, $\ba \in \RR^k$ denote the optimization variables, and $T\in M(d,d)$, $\bb \in \RR^d$ are fixed parameters. Recently, it was found \cite{arjevanifield2019spurious} that the symmetry of spurious minima 
of $\ploss$ \emph{break the symmetry} of global minima under various choices of data distributions (formal terms are given later). Here, for concreteness, we focus on the $d$-variate normal distribution for inputs
\cite{du2017gradient,zhong2017recovery,li2017convergence,tian2017analytical,
	ge2017learning,aubin2019committee}. 
	
\begin{remark}
In \cite{Arjevani2023deepsb}, the \emph{theory of distributions} is adopted as a central analytical framework for the rigorous treatment of the nonsmoothness inherent in ReLU networks. This approach applies to a broad class of data distributions given by Radon measures, that is, finite Borel measures of total mass one, and in particular includes the normal distribution as is the case in the present work. It furnishes a unified and mathematically precise foundation for differentiation with respect to network parameters, encompassing, in particular, bias terms (see the concluding remark in \pref{sec:concluding} below). Within this framework, techniques from \emph{geometric measure theory} provide a refined structural characterization of the gradient, which is viewed as a vector-valued function of bounded variation, as well as of the Hessian, now represented in the distributional sense as a matrix-valued Radon measure.
\end{remark}	
	
Using ideas based on symmetry breaking, techniques from representation theory and real algebraic geometry were employed to construct infinite families of critical points represented by Puiseux series in $d^{-1}$ and so obtain sharp estimates for the loss and the Hessian spectrum holding for \emph{finite}, arbitrarily large, dimensionality \cite{ArjevaniField2020,arjevanifield2020hessian,arjevanifield2021analytic}. We refer to \cite{arjevanifield2021tensor,arjevani2023symmetry} for similar analyses of tensor decomposition problems. The analytic results were used to investigate some of the key foundational phenomena occurring in DL. For example, the phenomenon of extremely skewed spectrum of the Hessian observed for large-scale trained networks \cite{bottou1991stochastic,lecun2012efficient,sagun2016eigenvalues,sagun2017empirical} was established, analytically, for families of minima and for arbitrarily large dimensionality. Other results concerned, for example, a long standing debate regarding whether some notion of local curvature can be used to explain generalization 
\cite{hochreiter1997flat, keskar2016large,jastrzkebski2017three,chaudhari2019entropy,dinh2017sharp}, ruling out, in the setting considered, notions of `flatness' relying exclusively on the Hessian spectrum. 

A particular phenomenon concerning two types of infinite families of critical points: type I and type II, was as follows. Despite type I and type II critical points being provably  local minima for (essentially) all $d\in\NN$, empirically, the former is never detected by standard gradient-based optimization methods initialized using, e.g., Xavier initialization. Thus, henceforth, type I minima shall also be referred to as \emph{hidden} minima. This favoring of type II minima over type I reflects a bias of optimization methods towards minima of reduced loss. Indeed, while the loss at type II minima converges to zero as $d$ increases ($O(1/d)$ for all known type II families), at type I minima the loss remains bounded away from zero. In particular, any hidden minimum is necessarily spurious (but not vice verse. Type II spurious minima exist and are detected by gradient-based methods). Of course, global minima of $\ploss$, loss zero, are type II.

\renewcommand{\arraystretch}{1.25}

\begin{table}[ht]\label{table:droot_spec}
\begin{center}
\begin{tabular}{c|ccccccccc}
	Eigenvalue &$\frac{\pi-2}{4\pi}$&$\frac{1}{4}$&$\frac{\pi+2}{4\pi}$&
	$\frac{d}{2 \pi} + \frac{- \pi^{2} - 4 + 6 \pi}{4 \pi \left(2 - \pi\right)} $&$\frac{d}{4} + \frac{- \pi^{2} - 2 \pi + 4}{4 \pi \left(2 - \pi\right)}$&$\frac{d}{4} + \frac{1}{4}$\\\hline
	Multiplicity&$\frac{d(d-1)}{2}$&$d-1$&$\frac{d(d-3)}{2}$&$1$&$1$&$d-1$
\end{tabular}
\end{center}
\caption{To $O(d^{-1/2})$-order, the Hessian spectrum at  type I and type II minima is~identical.}
\end{table}

\renewcommand{\fC}{C}
Attempting to argue about distinctive analytic properties of hidden minima using the loss Hessian spectrum, one finds that eigenvalues for both types 
agree modulo $O(d^{-1/2})$-terms, see \pref{table:droot_spec}. In addition, in both cases the expected loss at initialization is at least an order of magnitude larger than the loss at type I and type II minima. Our investigation thus proceeds by a technique introduced in \cite{ArjevaniCP2023}. Given a critical point $\fC\in M(k, d)$, $d$ and $k$ fixed, we consider the functions 
\begin{align} \label{prob:min}
m(r) &\defeq \min\{\ploss(W)~|~W\in \sphere_\fC(r)\},\\
\label{prob:max}
M(r) &\defeq \max\{\ploss(W)~|~W\in \sphere_\fC(r)\},
\end{align}
describing the minimum (resp. maximum) of $\ploss$ on $\sphere_\fC(r)$, the sphere of radius $r$ centered at $\fC$, Frobenius norm on $M(k,d)$. Of course, $m(r)$ and $M(r)$ are well-defined as $\sphere_\fC(r)$ is compact. Our approach to the pr	oblem of identifying distinctive properties of hidden minima proceeds by studying various aspects of arcs $\Gamma:[0,1)\to M(k,d)$ giving $m(r)$ and $M(r)$, in particular their structure and symmetry---the focus of this work.

A formal discussion of our results requires some familiarity with the representation theory of groups and o-minimal theory. Here, we provide high-level description of our contributions, briefly covering basic definitions from group theory, and defer more detailed statements to later sections after the relevant notions have been introduced (\pref{sec:rep}). We consider the natural (orthogonal) action of $S_k\times S_d$ on the parameter space $M(k,d)$: the first factor permutes rows, the second columns. Given a weight matrix $W\in M(k,d)$, the largest subgroup of $S_k\times S_d$ fixing $W$, the \emph{isotropy group} of $W$, is used as a means of measuring the symmetry of $W$. For example, the isotropy group  of the identity matrix $I_d$ is the diagonal subgroup $\Delta S_d \defeq \{(\pi,\pi)~|~\pi\in S_d\}\subseteq S_d\times S_d$. When $T$ possesses certain invariance properties, isotropy groups occurring for minima detected empirically are seen to be \emph{symmetry breaking} in the sense that they form \emph{proper} subgroups of $\Delta S_d$. 
In this work, we show that arcs minimizing \pref{prob:min} or maximizing \pref{prob:max} the loss emanating from symmetry breaking critical points are themselves---symmetry breaking. More specifically,

\begin{itemize}[leftmargin=*]
	
\item We give a detailed description of the possible intersections of subspaces invariant to subgroups of $S_d$ (the associated \emph{isotypic components}) with subspaces that are fixed by the action (\pref{thm:max_iso}). As tangency arcs of \emph{o-minimal definable} functions must approach critical points tangentially to Hessian eigenspaces (\pref{lem:tang_eigsp}), the latter amounts to obtaining an enumeration of all (generically finitely many) admissible structures and isotropy types for curves along which $C^2$ invariant functions are minimized and maximized. The methods apply beyond the groups considered in this work and so are of independent interest.

\item The general results are illustrated for 4 infinite families of minima of $\ploss$: $C_{p}^X$, with $X\in\{\mathrm{I},\mathrm{II}\}$ denoting the family type and $p\in\{0,1\}$ isotropy $\Delta (S_{d-p}\times S_p)$ (\pref{thm:type_iso}). We compute, with some effort, the two leading terms of the Hessian eigenvalues, displayed in \pref{table:sdTsdevs}, and show that it is precisely by these terms that the structure and symmetry of $m(r)$ associated with $\ploss$ are determined---differing, indeed, for type I and type II minima. The results cannot be obtained using existing analyses as the Hessian spectrum for type I and type II are identical to the order to which eigenvalue terms have been previously computed, see \pref{table:droot_spec}.

\item The general results are stated and proved for \emph{o-minimal structures}, simplifying and generalizing existing ones for symmetry breaking. For example, the existence and construction of Puiseux series describing families of critical points follow by a direct consequence of an o-minimal version of the Curve Selection Lemma for globally subanalytic~sets. 

\item Curves giving $m(r)$ and $M(r)$ are instances of symmetric tangency arcs developed in \cite{ArjevaniCP2023} to show that critical points connected to symmetric ones are symmetry breaking, see concluding remarks. Here, fundamental results from o-minimal theory  enable a numerical construction of tangency arcs, the continuation of which yields estimates on distances from minima, both types, to adjacent critical points. The numerical estimates conform with the theoretical analysis of the symmetry of $m(r)$ and $M(r)$.

\end{itemize}

\begin{table}[ht]
	\begin{center}
		\begin{tabular}{|l|l||c|c|c|c| }\cline{3-6}
			\multicolumn{2}{l||}{}
			&		
			$ \fC^{\mathrm{I}}_0$&$\fC_0^{\mathrm{II}}$ & \hspace*{0.05in}$\fC_1^{\mathrm{I}}$& $\fC_1^{\mathrm{II}}$\\\hline\hline
			\multicolumn{2}{|l||}{Type}&I&II&I&II\\\hline
			\multicolumn{2}{|l||}{Loss}&		
			$- \frac{1}{\pi} + \frac{1}{2} - \frac{4}{3 \pi \sqrt{d}}$&$0$&  $- \frac{1}{\pi} + \frac{1}{2} - \frac{4}{3 \pi \sqrt{d}}$&$\frac{-4 + \pi^{2}}{2 \pi^{2} d}- \frac{32}{3 \pi^{4} d^{\frac{3}{2}}}$\\\hline
			\multicolumn{2}{|l||}{Isotropy}&		
			$\Delta S_d$ &$\Delta S_d$ & $\Delta S_{d-1}$ & $\Delta S_{d-1}$\\\hline
			\multicolumn{2}{|l||}{Orbit length} &   $d!$& $d!$&$d\cdot d!$&$d\cdot d!$\\\hline
			Rep.  &Mult.
			\hspace*{0.05in}\\\hline
			$\ft$& $1$ & 
			&&$\frac{\pi -2}{4 \pi} + \frac{4 \left(-1 + \pi\right)}{\pi^{3} d}$& $\frac{\pi -2}{4 \pi} + \frac{2 \left(\pi -2\right)}{\pi^{2} d}
			$\\
			&&&&$\frac{1}{4} + \frac{- 50 \pi + 24 + \pi^{3} + 10 \pi^{2}}{\pi^{4} d^{2}}$ &$\frac{1}{4} + \frac{-1 + 2 \pi}{\pi^{2} d}
			$\\&&$\frac{d}{2 \pi} + \frac{- \pi^{2} - 4 + 6 \pi}{4 \pi \left(2 - \pi\right)} $&$\frac{d}{2 \pi} + \frac{- \pi^{2} - 4 + 6 \pi}{4 \pi \left(2 - \pi\right)} $&$\frac{d}{2 \pi} + \frac{- \pi^{2} - 4 + 6 \pi}{4 \pi \left(2 - \pi\right)}$&$\frac{d}{2 \pi} + \frac{- \pi^{2} - 4 + 6 \pi}{4 \pi \left(2 - \pi\right)}$\\
			&&&&$\frac{d}{4} + \frac{1}{4}$&$\frac{d}{4} + \frac{1}{4}$\\
			&&$\frac{d}{4} + \frac{- \pi^{2} - 2 \pi + 4}{4 \pi \left(2 - \pi\right)}$&$\frac{d}{4} + \frac{- \pi^{2} - 2 \pi + 4}{4 \pi \left(2 - \pi\right)}$&$\frac{d}{4} + \frac{- \pi^{2} - 2 \pi + 4}{4 \pi \left(2 - \pi\right)}$&$\frac{d}{4} + \frac{- \pi^{2} - 2 \pi + 4}{4 \pi \left(2 - \pi\right)}$\\
			\hline
			$\fs$ & $d-p-1$ &$\frac{\pi -2}{4 \pi}$&\colorbox{myblue}{$\frac{\pi -2}{4 \pi}$}& $\frac{\pi -2}{4 \pi} + \frac{2 - \pi}{2 \pi^{2} d}$ &
			\colorbox{myblue}{
				$\frac{\pi -2}{4 \pi} - \frac{1}{\pi^{2} \sqrt{d}}$} \\
			&&&&$\frac{\pi -2}{4 \pi}$&$\frac{\pi -2}{4 \pi} + \frac{- \frac{\pi^{3}}{2} - 8 - \pi}{\pi^{4} d}$\\
			&&$\frac{1}{4} - \frac{2}{\pi \sqrt{d}}$&$\frac{1}{4} + \frac{-1 + \pi}{\pi^{2} d}$& $\frac{1}{4} + \frac{-3 + 2 \pi}{\pi^{2} d}$&$\frac{1}{4} + \frac{- 2 \pi^{2} - 8 + 7 \pi}{\pi^{3} d}$\\
			&&&&$\frac{\pi + 2}{4 \pi} + \frac{3 \cdot \left(2 - \pi\right)}{2 \pi^{2} d}$&$\frac{\pi + 2}{4 \pi} - \frac{1}{\pi^{2} \sqrt{d}}$\\
			&&$\frac{d}{4} + \frac{1}{4}$&$\frac{d}{4} + \frac{1}{4}$&$\frac{d}{4} + \frac{1}{4}$&$\frac{d}{4} + \frac{1}{4}$\\
			\hline
			$\fx$ & $\frac{(d-p-1)(d-p-2)}{2}$ & \colorbox{myblue}{$\frac{\pi -2}{4 \pi} - \frac{1}{\pi \sqrt{d}}$} &$\frac{\pi -2}{4 \pi}$&	\colorbox{myblue}{$\frac{\pi -2}{4 \pi} - \frac{1}{\pi \sqrt{d}}$} &$\frac{\pi -2}{4 \pi} - \frac{1}{\pi d}$ \\
			\hline
			$\fy$ & $\frac{(d-p)(d-p-3)}{2}$ &$\frac{\pi + 2}{4 \pi} - \frac{1}{\pi \sqrt{d}}$ &$\frac{\pi + 2}{4 \pi}$	&$\frac{\pi + 2}{4 \pi} - \frac{1}{\pi \sqrt{d}}$&$\frac{\pi + 2}{4 \pi}$  \\
			\hline
		\end{tabular}
	\end{center}
		\caption{Dominating terms of the Puiseux series describing the loss and the Hessian spectrum for 4 families of minima. The structure and symmetry of curves along which the loss is minimized is determined by the irreducible	representations of $S_d$ with which the minimal eigenvalue (highlighted) is associated. 
		For both types, the maximal eigenvalues (having leading terms growing linearly with $d$)  belong to the $\ft$- and $\fs$-representation, implying that the dynamics concentrates in the vicinity of small subspaces of multiplicity~$O(d)$.}
	\label{table:sdTsdevs}
\end{table}

\section{Framework: the tangency set and symmetry}
\label{sec:tanset}

Focusing on the quantities $m(r)$ \pref{prob:min} and $M(r)$ \pref{prob:max} as a means of investigating hidden minima, one is naturally led to a more general consideration of curves describing critical points of $\ploss|_{\sphere_\c(r)}$, or equivalently curves lying in a set, the \emph{tangency set} defined below, comprising all points where level sets of $f$ lie \emph{tangential} to spheres centered at~$\c$. The tangency set arises naturally in the study of singularities, for examples  \cite{netto1984jet,nemethi1992milnor,le1998bifurcation,durfee1998index,pham2016genericity}. In this work we show that a tangency set inherits symmetries from  the function with which it is associated, a result which we then use for characterizing hidden minima (\cite{ArjevaniCP2023}). A formal discussion of our main results requires some familiarity with group, representation and O-minimal theory, which we briefly review below. Proofs are deferred to the appendix.

\begin{definition}
Suppose given a $C^1$ function $f:\RR^d\to\RR$ and 
a point $\c\in \RR^d$. 
\begin{itemize}[leftmargin=*] 
\item The set of critical points of $f$ is $\critset(f) \defeq \{\x\in \RR^d~|~D f(\x) = 0\}$.
\item The \emph{tangency set} $\tanset_\c(f)$ relative to $\c$ is defined by,
\begin{align}
\tanset_c(f) \defeq \{ \x \in\RR^d~|~ D_if(\x)\x_j = D_jf(\x)\x_i,~i,j\in [d]\}.
\end{align}
In particular, $\critset(f)\subseteq \tanset_c(f)$. 
\item 
A \emph{tangency arc} relative to $\c$ is a $C^1$-embedding $\gamma:[0,1)\to \RR^d$ satisfying $\gamma(0) = \c$ and $\gamma(t) \in \tanset_\c\backslash\{\c\}$ for 
$t\in(0,1)$. We say that $\gamma$ is parameterized by arc length if $\|\gamma(t)-\c\|=t,~t\in[0,1)$, $\|\cdot\|$ denoting the standard Euclidean norm throughout.  
\end{itemize}
\end{definition}

We typically do not indicate the dependence  of $\critset$ and $\tanset_\c$ on $f$ if no confusion results. 

The tangency set has a particularly simple structure for quadratic forms. If $f:\RR^d\to\RR$ is given by  $f(\x) = \x^\top A \x/2$, $A$ symmetric, then relative to $\c=0$,
\begin{equation}\label{eqn:quad_tan}
\tanset_0 = \{\x \in \RR^d~|~\exists \eta\in\RR, A\x= \eta \x, ~\x\neq 0\},
\end{equation}
the set of all eigenvectors (see \pref{fig:hc_tan} for another example).
\newcommand{\minimizers}{X_\c^m} 

In \cite{ArjevaniCP2023}, the following general result is established, relying on methods developed for the study of bifurcation phenomena in variational problems:\\
\noindent\textbf{Theorem} (Informal) If $\c$ is an isolated critical point, then, \emph{generically}, \\
1. there are finitely many tangency arcs, each tangent to a Hessian eigenspace.\\
2. For every maximal isotropy subgroup $G \subset \Gamma$, there exists a tangency arc with isotropy $G$; if $\dim(V^\Gamma) > 0$, there exists an arc with isotropy~$\Gamma$.\\
3. the index of the arc, defined to be the number of eigenvalues below 
$\inner{\nabla f(\x), \x-\c}/\|\x-\c\|^2$, is constant along the (open) arc. In particular, at critical points of $f$, the index of the arc and that of the Hessian~coincide.\\

To maintain the focus of the manuscript, we do not develop the full theory presented in \cite{ArjevaniCP2023}. Instead, we restrict attention to the result above and note that a more extensive analysis is possible both globally, using for example topological methods \cite{rabinowitz1971some}, and locally, by incorporating higher-order derivatives to determine, for instance, the number of tangency arcs, their isotropy groups (not all of which need be maximal), and their indices. Below we present the results required for this paper, formulated within the framework of o-minimal structures.

\subsubsection{O-minimal theory}
The topology of tangency sets so defined can be quite complicated for arbitrary $C^1$ functions. However, for our applications, certain structural restrictions apply: the loss function $\ploss$, as we show, is \emph{definable} in an \textit{o-minimal structure} expanding the real field  \cite{van1998tame,van1996geometric}, that is, a sequence $\cD = (\cD_n)_{n\in\NN}$, $\cD_n$ denoting a collection of subsets of $\RR^n$, satisfying certain axioms which we detail in \pref{sec:omin}. Here, suffices it to note that sets defined by first-order formulae ranging over \emph{definable sets}, i.e., sets belonging to $\cD$, are themselves definable. A map $F:A\to\RR^n$ is called \emph{definable} if its graph  $\Gamma(F) = \{(x, F(x))~|~ x\in A\} $ is definable. Thus, if $f:\RR^d\to\RR$ is definable, the set $\minimizers \defeq \cup_{r>0} \argmin f|_{\sphere_\c(r)}$ consisting of all points at which $m(r)= \min\{f(\x)~|~\x\in\sphere_r(\c)\}$ (abusing notation in (\ref{prob:min})) is attained, may be equivalently given by 
\begin{align}\label{eqn:logic_minimizers}
	\{\x\in\RR^d~|~\exists r\in \RR, \forall \y\in\RR^d (\|\y-\x\|^2 = r^2 \implies \exists s\in\RR, f(\y) - f(\x) = s^2)   \}
\end{align}
and so is definable (leaving the validation of use of abbreviations such as $f(\x)-f(\y) = s^2$ to the reader). Throughout, all sets and mappings involved in our analysis are definable, assuming $f$ is, as we may as $\ploss$ is (most of what we do also applies to smooth stratified mappings \cite{mather1973stratifications}). The proofs are straightforward and so are omitted. A prototypical example of an o-minimal structure is given by semi-algebraic sets. The loss function $\ploss$ is definable in the larger o-minimal structure $\RR_{an}$ of \emph{globally subanalytic sets} given by inverse images of subanalytic sets \cite{lojasiewicz1995semi} under the Nash (algebraic and real analytic) map $\cV_d(\x) \defeq \prn*{{x_1}/{\sqrt{1+\|\x\|^2}},\dots,{x_d}/{\sqrt{1+\|\x\|^2}}}$ mapping $\RR^{d}$ isomorphically onto $(-1,1)^d$, see \pref{sec:omin}.

O-minimal structures, motivated as a candidate for Grothendieck's idea of `tame topology' \cite{grothendieck1997esquisse}, offer a framework flexible enough to carry out geometrical and topological constructions on real Euclidean set (e.g., projection) and real functions (e.g., composition and differentiation), yet sufficiently restrictive to impose certain regularity, as demonstrated by the following o-minimal theoretic result (or a metric version thereof).\\

\paragraph{Curve Selection Lemma (CSL)} 
If $a\in \cl{X}$, where $X$ is definable, then there exists a definable continuous map $\gamma:[0,1)\to X$ such that $\gamma((0,1))\subseteq X$ and $\nrm{\gamma(r) - a} = r$. \\

\noindent By a direct Lagrangian multipliers argument, $\c$ lies in the closure of $\minimizers$ (definable, by (\ref{eqn:logic_minimizers})). Thus, by the CSL, 

\begin{corollary} \label{corr:mr_exists}
	If $f$ is definable then there exists a tangency arc $\gamma$ parameterized by arc length satisfying $\ploss(\gamma(r)) = m(r)$, and similarly for $M(r)$.
\end{corollary}
\noindent Despite the simplicity of the proof, the existence of an arc giving $m(r)$ is certainly not obvious. Counter-examples exist already for (necessarily non-definable) functions on $\RR$, e.g., $x\mapsto \sigma^7(x)\sin(1/x)$. For quadratic functions, see (\ref{eqn:quad_tan}), a tangency arc giving $m_d(r)$ must lie in the eigenspace associated to the minimal eigenvalue, therefore so does $\dot{\gamma}(0)$, if exists. More generally, we have the following lemma,
\begin{lemma} \label{lem:tang_eigsp} 
	Suppose given a $C^2$ definable function $f:U\to\RR$, $U$ open, and a critical point $\c\in U$. Any tangency arc parameterized by arc length approaching $\c$ must do so tangentially to an eigenspace of $\nabla^2 f(\c)$ (in particular, $\dot\gamma(0)$, as a one-sided limit, exists). Moreover, tangency arcs giving $m(r)$ (resp. $M(r)$) are tangential to the eigenspace associated to the minimal (resp. maximal) eigenvalue.
\end{lemma}
The lemma is proved using standard results from perturbation theory and the \emph{monotonicity theorem}, an elementary result in o-minimal theory. The correspondence stated between tangency arcs and eigenspaces (see \pref{fig:hc_tan} in the concluding remarks for illustration) becomes particularly useful for invariant functions on account of the detailed information on Hessian invariant subspaces holding, a-priori, for all points of fixed symmetry.

\subsubsection{Representation theory of groups} \label{sec:rep}
We give a brief, if terse, review of group and representation theory that suffices for our applications. For more detail and generality see \cite{thomas2004representations}.

Given a vector space $V$, a group $G$ and a point $\x \in V$, the largest subgroup of $G$ fixing $\x$ is called the \emph{isotropy} subgroup of $\x$ and is denoted by $G_\x$. We let $(G_\x)$ denote the conjugacy class of the subgroup $G_\x$ in $G$ and say that $(G_\x)$ is the \emph{isotropy type} of $\x$. The set of isotropy types inherits the partial order given for the lattice of subgroups with the partial order relation being set inclusion. If $G$ is a subgroup of $\Od{\real^d}$, the action on $\real^d$ is called an \emph{orthogonal} representation of $G$ (we often drop the qualifier orthogonal). The \emph{symmetric group} $S_d$, $d\in\pint$, is the group of permutations of $[d]\defoo \{1,\dots,d\}$. We may identify $S_d$ with the subgroup of $\text{O}(\real^d)$ consisting of permutation matrices. Thus $S_d$ acts orthogonally on $\real^d$ (as permutation matrices), as does $S_k \times S_d\subset S_{k\times d}$ on $M(k,d)$ (first factor permuting rows, second factor columns) with respect to the standard Euclidean inner product on $M(k,d) \approx \real^{k\times d}$. The \emph{degree} of a representation $(V,G)$ is the dimension of $V$. Given two representations $(V_1,G)$ and $(V_2,G)$, a map $A: V_1 \arr V_2$ is called $G$-equivariant if $A(gv) = gA(v)$, for all $g \in G, v \in V_1$. If $A$ is linear and equivariant, we say $A$ is a \emph{$G$-map}. When $G$ acts \emph{trivially} on $V_2$, that is $g\x=\x$ for all $g\in G$ and $\x\in V_1$, $A$ is said to be invariant. If $H$ is a group then one key feature of $H$-invariant differentiable functions is the $H$-equivariance of their gradient fields. These are naturally expressed in terms of fixed point linear subspaces  
defined by $V^G \defoo \{y \in V \dd hy =y, \forall g\in G\},~ G\subseteq 
H$. Thus if $f$ is $H$-invariant, the gradient $\nabla F$ is a $H$-equivariant self map of $M(k,d)$, and if $\c$ is a critical point of $\nabla F$ with isotropy $G\subseteq H$ then $\nabla^2 f: M(k,d)\arr M(k,d)$ is a $G$-map.
For sequences of groups $(G_d)$ and target matrices $(T_d)$ considered in this work, $k-d$ fixed, inner products between rows of $W\in M(k, d)^{G_d}$ and $T_d$ may be expressed in terms of polynomials in $d$ and $N$ variables by identifying $M(k, d)^{G_d}$ with $\RR^N$ by a suitable linear isomorphism $\Xi\defeq \Xi(d):\RR^N\to M(k,d)^{G_d}$ ($d$ sufficiently large). We call such sequences \emph{natural}. A~representation $(V, G)$ is \emph{irreducible} if the only linear subspaces of $\real^n$ that are preserved (invariant)  by the $G$-action are $\real^n$ and $\{0\}$. Two orthogonal representations $(V_1,G)$, $(V_2,G)$ are \emph{isomorphic} if there exists a $G$-map $A: V_1 \arr V_2$ which is a linear isomorphism. If $(V_1,G)$, $(V_2,G)$ are irreducible but not isomorphic then every $G$-map $A: V_1 \arr V_2$ is zero (as the kernel and the image of a $G$-map are $G$-invariant). If $(V,G)$ is irreducible, then the space $\text{Hom}_G(V,V)$ of $G$-maps (endomorphisms) of $V$ is a real associative division algebra and is isomorphic by a theorem of Frobenius to either $\real, \mathbb{C}$ or $\mathbb{H}$ (the quaternions). The \emph{only} case that will concern us here is when $\text{Hom}_G(V,V) \approx \real$ when we say the representation is \emph{absolutely irreducible}. Every representation $(V, G)$ can be written uniquely, up to order, as an orthogonal direct sum $\oplus_{i\in\ibr{m}} V_i$, where each $(V_i,G)$ is an orthogonal direct sum of isomorphic irreducible representations $(V_{ij},G)$, $j \in \ibr{p_i}$, and $(V_{ij},G)$ is isomorphic to $(V_{i'j'},G)$ if and only if $i' = i$. Although \emph{not} uniquely determined if $p_i > 1$, a judicious choice of $V_{ij}$ and representative vectors yields a complete derivation of all eigenvalues of $G$-maps. The technique is standard (see \cite{GSS1998} for a more general account) and is used in the present work to derive, with some effort, $O(d^{-1/2})$-eigenvalue terms (see \pref{table:sdTsdevs}). If there are $m$ distinct isomorphism classes $\mathfrak{v}_1,\cdots,\mathfrak{v}_m$ of irreducible representations, then $(V, G)$ may be represented by the sum $p_1 \mathfrak{v}_1 + \cdots + p_m \mathfrak{v}_m$, where $p_i\ge 1$ counts the number of representations with isomorphism class $\mathfrak{v}_i$. Up to order, this sum (that is, the $\mathfrak{v}_i$ and their multiplicities) \emph{is} uniquely determined and is called the \emph{isotypic decomposition} of $(V, G)$. In \pref{sec: iso_global} we describe our use of the isotypic decomposition for arguing about the symmetry of tangency arcs using  \pref{lem:tang_eigsp}---only assuming $S_k\times S_d$-invariance. 

\subsubsection{Expressing curves of critical points as Puiseux series in $d$} 
The loss function $\ploss$ is not only structurally tame, being $\RR_{an}$-definable, but also has an important geometric characteristic: functionally, it only depends on inner products between the rows of $W$ and $T$, disregarding for the present weights belonging to the second layer to make the arguments below more transparent. Thus, for natural sequences (notation as in the preceding section), we may express $\ploss$ as a (definable) function of polynomials in $d$ and $\bxi\in\RR^N$, and so regard $d$ as a real variable and so have the set $\sigma_\ploss \defeq \{(\bxi, d)\in \RR^{N+1} ~|~\Xi(\bxi)\in \critset(\ploss(\cdot;d)),~d\ge d_0\}$ $\RR_{an}$-definable, $d_0\in\NN$ suitably chosen. Each infinite sequence of critical points $W_d = \Xi(\bxi_d), d \in \{d_0,d_0+1,\dots\}$ and $(\bxi_d, d)\in\sigma_\ploss$
gives a point in $\cl{\cV_{N+1}(\sigma_\ploss)} \cap \partial [-1,1]^{N+1}$ which, by the CSL, may be approached by a definable arc lying in $\cV_{N+1}(\sigma_\ploss)$, giving in turn a definable curve of critical points in $M(k,d)$. In $\RR_{an}$, the entries of the resulting curve are in fact Puiseux series (in $1/d$), the coefficients of which may be computed in an obvious gradual manner, see \pref{sec:puiseux}.

Another important implication of $\sigma_L$ being definable is that by 
\emph{local triviality} the topological type of $\Xi^{-1}\critset(\ploss(\cdot;d))$ is identical for sufficiently large $d$. A continuous definable map $f:E\to B$ between definable sets is \emph{definably trivial} if there are a definable set $F$, the  \emph{fiber}, and a definable homeomorphism $h:E\to B\times F$ such that $f\circ h^{-1} = \pi_1,~\pi_1:B\times F\to B$ being the projection on the first factor. Definable maps are locally trivial in the sense that the set $B$ can partitioned into definable sets $B_1,\dots,B_n$ such that $f|_{f^{-1}(B_i)}, i\in [n],$ are definably trivial. The assertion above now follows by considering the projection of $\sigma_L$ on the $d$-coordinate. In addition, we see that arcs of critical points may bifurcate, as they sometimes do, when $d$ is not sufficiently large.

\section{Main results: structure and symmetry of tangency~arcs} \label{sec: iso_global}
The isotropy groups that correspond to type I and type II minima $k = d$ (\pref{def: types} below) are \emph{diagonal} subgroups of the form $\Delta (S_{i_1}\times \cdots \times S_{i_q}) \defeq \{(g,g) \dd g \in S_{i_1}\times \cdots \times S_{i_p}\subseteq S_d\}\subseteq S_d\times S_d,~i_1+\cdots+i_q=d$. Clearly, $S_{i_1}\times \cdots \times S_{i_p} \cong  \Delta (S_{i_1}\times \cdots \times S_{i_q})$. To indicate how results given in this section are obtained, consider the orthogonal direct sum decomposition
/\begin{equation} \label{eqn:ort_dec}
M(d, d) = \bD_d \oplus \bS_d \oplus \bA_d,
\end{equation}
with $\bD_d$ denoting the space of diagonal $d\times d$ matrices, $\bA_d$ the space of skew-symmetric $d\times d$ matrices and $\bS_d$ the space of symmetric $d\times d$ matrices with diagonal entries zero. Since $S_d$ acts diagonally on $M(d, d)$, the factors are $S_d$-invariant. We may now ask: relative to the ascending series of subgroups of $\Delta S_d$, 
\begin{equation} \label{eqn:asc_groups}
1 = \Delta S_1^d \le \cdots  \le \Delta(S_2\times S_1^{d-2}) \le \Delta(S_{d-1}\times S_1) \le \Delta S_d,
\end{equation}
what is the maximal $i\in\{0,1, \dots,d\}$ such that $M(d, d)^{\Delta(S_i\times S_1^{d-i})}$ intersects a given factor in (\ref{eqn:ort_dec})? Computing, we see that whereas $M(d,d)^{\Delta S_d}\cap \bD_d = \xi I_d$ and $M(d,d)^{\Delta S_d}\cap \bS_d  = \xi (\bones \bones^\top - I_d)$, the third factor $\bA_d$ does not intersect $M(d,d)^{\Delta S_d}$. However, $\bA_d$ does intersect $M(d,d)^{\Delta (S_{d-1} \times S_1)}$. Thus, if $\bA_d$ were an eigenspace of a given linear transformation then, referring to (\ref{eqn:asc_groups}), any associated eigenvector would have isotropy type at most $\Delta (S_{d-1}\times S_1)$. For type I and type II points, $\bA_d$ is \emph{not} an eigenspace. However, a similar reasoning, involving isotypic components
rather than eigenspaces, applies. The isotypic decomposition of $(M(d, d),S_d)$ is relatively simple and uses just 4 irreducible representations of $S_d$, each associated to a partition of the set $[d]$: the trivial representation $\mathfrak{t}$ of degree 1 associated to the partition $(d)$, the standard representation $\mathfrak{s}_d$ of degree $d-1$ associated to $(d-1,1)$, the exterior square representation $\mathfrak{x}_d=\wedge^2 \mathfrak{s}_d$ of degree ${(d-1)(d-2)}/{2}$ associated to $(d-2,1,1)$, and a representation $\mathfrak{y}_d$ of degree ${d(d-3)}/{2}$ associated to $(d-2, 2)$.

\begin{theorem}\label{thm:max_iso}
Write $M(d,d) = \bV_\ft \oplus \bV_\fs \oplus \bV_\fx \oplus \bV_\fy $, with the factors respectively denoting the trivial, standard, exterior square and $\fy$ isotypic component. If $A$ is an $S_d$-map, then,
\begin{itemize}[leftmargin=*]
\item $\bV_\ft=M(d,d)^{\Delta S_d}$. In particular, referring to (\ref{eqn:asc_groups}), the maximal isotropy type of matrices in $\bV_\ft$ is $\Delta S_d$. The spectrum of $A|_{\bV_\ft}$ and $A|_{\bV_\ft\cap M(d,d)^{\Delta S_{d}}}$ are identical.

\item In $\bV_\fs$, the maximal isotropy type  is $\Delta (S_{d-1}\times S_1)$. Every eigenvalue of $A|_{\bV_\fs}$, necessarily of multiplicity $i(d-1), i\in[3]$, is an eigenvalue of $A|_{\bV_\fs\cap M(d,d)^{\Delta (S_{d-1}\times S_1)}}$ of multiplicity~$i$.

\item In $\bV_\fx$, the maximal isotropy type is $\Delta (S_{d-2}\times S_1^2)$. 
The map $A|_{\bV_\fx\cap M(d,d)^{\Delta (S_{d-2}\times S_1^2)}}$ has a single eigenvalue, multiplicity one, given by the (distinct) single eigenvalue of $A|_{\bV_\fx}$, multiplicity $(d-1)(d-2)/2$.

\item In $\bV_\fy$, the maximal isotropy type is $\Delta (S_{d-2}\times S_1^2)$. The map $A|_{\bV_\fy\cap M(d,d)^{\Delta (S_{d-2}\times S_1^2)}}$ has a single eigenvalue, multiplicity one, given by the single eigenvalue of $A|_{\bV_\fy}$, multiplicity $d(d-3)/2$.
\end{itemize}
A detailed description of the intersections of the subspaces involved is provided in the proof of the theorem in \pref{sec:thm_max_iso_prf}.
\end{theorem}
\newcommand{\spec}{\sigma}
Combined with \pref{lem:tang_eigsp}, we obtain the following general result quantifying the amount of symmetry breaking needed for realizing extremal tangency arcs.
\begin{corollary} (Notation and assumptions as above.)\label{corr:eig_sym}
Suppose $C\in M(d,d)$ is a critical point with isotropy $\Delta S_d$ of an $S_d$-invariant definable function, and $\gamma$ a tangency arc tangential to an $\mu$-eigenspace of $\nabla^2 f(C)$. If $\mu$ belongs to $\nabla^2 f(C)|_{\bV_\ft}$, then the maximal isotropy type of $\dot{\gamma}(0)$ is $\Delta S_d$. For $\nabla^2 f(C)|_{\bV_\fs}$ (resp. $\nabla^2 f(C)|_{\bV_\fx}$ or $\nabla^2 f(C)|_{\bV_\fy}$), the maximal isotropy type is $\Delta (S_{d-1}\times S_1)$ (resp. $\Delta (S_{d-2}\times S_1^2)$).	
\end{corollary}
The related measure quantifying the \emph{minimal} isotropy type occurring for a given isotypic component plays a major role in the study of non-local aspects of symmetric tangency sets concerning, in particular, symmetry of critical points \cite{ArjevaniCP2023}, and see concluding remarks. 

\subsection{Structure and symmetry of tangency arcs of type I and type II minima} \label{sec: general_to_types}
We now apply the general results to type I and II minima. 
\begin{definition}\label{def: types}
Let $p \ge 0$ and take $G_d = \Delta(S_{d-p} \times S_p)$. A family of critical points with isotropy $(G_d)_{d \ge d_0}$ is type I (resp.~type II)
if as $d \arr \infty$, the diagonal elements of the $(d-p)\times 
(d-p)$-block corresponding to the action of $\Delta 
S_{d-p}$ converge to $-1$ (resp.~$+1$).
\end{definition}
Below, we emphasize families with isotropy $\Delta(S_{d-p}\times S_p),~ p\in \{0,1\}$ and $k=d$. Methods and results apply to other values of $p$, as well as for $k>d$. By \pref{corr:eig_sym}, adapted to $\Delta (S_{d-p}\times S_p)$-maps, we have,

\begin{theorem}\label{thm:type_iso}
The Hessian eigenvalues of type I and type II minima isotropy $\Delta(S_{d-p}\times S_p), p\in \{0,1\}$ represented by Puiseux series computed to two leading terms is given in \pref{table:sdTsdevs}. In particular, identifying the isotypic components giving the minimal and maximal eigenvalues, we find that,
\begin{enumerate}[label=\Alph*., leftmargin=*]
\item The maximal isotropy type of  tangency arcs giving $m(r)$ for type I (resp. II) is $\Delta (S_{d-p-2}\times S_{p+2})$ (resp. $\Delta (S_{d-p-1}\times S_{p+1})$).
\label{thm:type_iso_a}
\item The maximal isotropy type of tangency arcs giving $M(r)$ is at least $\Delta (S_{d-p}\times S_{p})$. 
\label{thm:type_iso_b}
\end{enumerate}
\end{theorem}
Assertions A and B, obtained by pure group representation-theoretic considerations, are easily verified by considering small values of $r>0$ using simple numerical procedures such as projected gradient descent. In the next section larger values of $r$ for $m(r)$ are considered. We note that minimal eigenvalues are also responsible for determining the isotypic component along which minima are created \cite{arjevanifield2021analytic,arjevanifield2022equivariant}.

\renewcommand{\epsilon}{\varepsilon}
\section{Numerical results: bounding the distance to the nearest critical point} \label{sec:bound}
Our analysis thus far concerned local aspects of tangency arcs, specifically their symmetry and structure in the vicinity of a critical point $\c$. However, tangency arcs can be extended until a singularity is hit and so detect adjacent critical points. In fact, since the tangency set is definable, local triviality implies that \emph{any} critical point sufficiently close to $\c$ may be connected to by a tangency arc.  As with the metric version of the CSL, we consider the Euclidean norm $\|\cdot-\c\|:\tanset_\c\to \RR$. By local triviality, there exists an interval $(0,\epsilon)$, a definable set $F$ and a definable homeomorphism $h:\tanset_\c\cap \mathring{B_\epsilon}(\c)\to (0,\epsilon)\times F$. Recalling that $\critset\subseteq\tanset_\c$, if $\c'$ is a critical point at distance at most $\epsilon$ from $\c$ then $\c' = h^{-1}(\epsilon',\x')$ for some $\epsilon'\in(0,\epsilon)$ and $\x'\in F$. We may now define a tangency arc $\gamma(t) = h^{-1}(t\epsilon', \x)$ $(\gamma(0) \defeq \c)$ connecting $\c$ to $\c'$. The set $F$ being definable has a finite number of definable (in fact, piecewise $C^1$-path) connected components. As a simple consequence, in each component, $t\mapsto f(\gamma(t)), t\ge0,$ is identical for any tangency arc parameterized by arc length, whence regarding the number of tangency arcs being essentially finite (see \cite{netto1984jet} for similar results for germs of real polynomials). All of this indicates a simple effective means of estimating distances to critical points adjacent to $\c$: construct tangency arcs parameterized by arc length, numerically. We note that tangency arcs may also be given by analytic expressions, as with tensor decomposition problems \cite{arjevani2023symmetry}.

In practice, we construct tangency arcs for $\tanset_C(\ploss)$, $C\in M(d,d)$ a critical point, using a Lagrangian function encoding a norm constraint,
$Q(W,\eta, r) = \ploss(W) + \eta(\|W-C\|^2 - r^2)$.
The (augmented) tangency arcs $(W(r),\eta(r))$ are computed by solving $DQ = 0 $ using Newton-Raphson method and small increments of $r$, typically 1e-3, until hitting a singular Jacobian or until $r_{\text{max}}$ has been reached. The former case indicates the presence of a critical point or an arc bifurcating (both occur). The latter case is taken to indicate our (very) finite approximation for an arc continuing indefinitely. The limit $r_{\text{max}}$ has been chosen so as to be an order of magnitude larger than the typical length of arcs terminating at `finite' time. Following \pref{lem:tang_eigsp}, arcs are initialized by setting $W(0) \defeq C + r_{\text{min}}B$ where $B$ denotes an eigenvector and $r_{\text{min}}=10^{-7}$. Repeating this process for different minima and ranges of eigenvalues, we found that arcs corresponding to eigenvectors associated to small eigenvalues tend to be finite and generally terminate earlier than those corresponding to large eigenvalues. In \pref{table:arc_lengths}, we report the radius of arcs corresponding to minimal eigenvalues. The numerical estimates are consistent with \pref{thm:type_iso}: 
\begin{enumerate}[leftmargin=*]
\item For $C_0^I$ and $C_1^{II}$, the symmetry of an arc tangency realizing $m(r)$ is $\Delta (S_{d-2}\times S_1^2)$ (\pref{thm:type_iso}.I.) and so when the ambient space $M(d,d)^{\Delta (S_{d-1}\times S_1)}$ is replaced with the larger space $M(d,d)^{\Delta (S_{d-2}\times S_1^2)}$, radii drop as expected.

\item For $C_1^{I}$, the symmetry of an arc tangency realizing $m(r)$ is $\Delta (S_{d-3}\times S_1^3)$ (\pref{thm:type_iso}.II.) and so replacing the ambient space $M(d,d)^{\Delta (S_{d-1}\times S_1)}$ with the larger space $M(d,d)^{\Delta (S_{d-2}\times S_1^2)}$ shows no significant change. At $M(d,d)^{\Delta (S_{d-3}\times S_1^3)}$, radii drop.
\end{enumerate}
The results provided also suggest that for small $d$ the distance to the nearest critical point is small for spurious minima ($C_0^{I}, C_1^{I}$ and $C_1^{II}$) compared to that of the global minima $C_0^{II}$. The situation changes when $d$ increases. Additional phenomena arising from the numerical study of the tangency set are a topic of current work. Non-local aspects concerning symmetry of critical points as well as tangency arcs of minimal isotropy, are studied in some depth in \cite{ArjevaniCP2023}.

\begin{table}[h]\label{table:arc_lengths}
\begin{center}
\begin{tabular}{c|c|c|c|c|c|c|c|c}
Ambient space&$d$&$C_0^I$&$C_0^{II}$&$C_1^I$&$C_1^{II}$
\\\hline
\multirow{3}{*}{$M(d,d)^{\Delta (S_{d-1}\times S_1)}$} 
&7& 1.16 & 1.16 & 1.25 & 0.90 
\\\cline{2-2}
&20&1.05 & 1.05 & 1.1 & 0.90 
 \\\cline{2-2}
&100&1.01 & 1.01 & 1.03 & 0.97   \\\hline
\multirow{3}{*}{$M(d,d)^{\Delta (S_{d-2}\times S_1^2)}$} 
&7&
\colorbox{myblue}{0.62} & 1.75 & 1.12 & \colorbox{myblue}{0.31}
 \\\cline{2-2}
&20&1.01 & 1.53 & 1.08 & 0.81  \\\cline{2-2}
&100&1.05 & 1.01 & 1.05 & 0.99  \\\hline
\multirow{3}{*}{$M(d,d)^{\Delta (S_{d-3}\times S_1^3)}$} 
&7& 0.62 & $\infty$ & \colorbox{myblue}{0.29} & 0.31   \\\cline{2-2}
&20&1.01 & 1.45 & 1.01 & 0.81   \\\cline{2-2}
&100&1.42 &  & 1.05 & 0.99   \\\hline
\end{tabular}
\end{center}
\end{table}

\section{Concluding remarks and future work} \label{sec:concluding}

The focus in this paper has been on $m(r)$ and $M(r)$, giving respectively the minimal and maximal value of the loss function in the vicinity of critical points. The functions are studied as a means of identifying analytic properties differentiating hidden minima, type I, from ones detected by standard gradient-based methods, type~II. We prove general results on tangency arcs realizing $m(r)$ and $M(r)$ showing how pure group representation-theoretic considerations yield a precise description of the admissible types of such arcs. The general results used for the loss function $\ploss$ reveal that tangency arcs of type I differ from type II by their structure and symmetry, provably requiring a greater extent of \emph{symmetry breaking} to have curves realizing $m(r)$ on account of $O(d^{-1/2})$-terms of the Hessian spectrum of type I and II which are otherwise identical. The theoretical results are derived using methods developed for o-minimal structures which imply in particular a certain topological regularity required for the numerical work presented. In addition to confirming the structure and symmetry type predicted by our theoretical results, the construction of tangency arcs provides an effective means for studying critical points adjacent to a given one.

\subsubsection{Bias terms}

A natural question arising from the analysis of minima in this work is how the conclusions drawn in the unbiased setting are affected by the introduction of bias terms. The biased analogue of the kernel appearing in the unbiased analysis is as follows. For $\mathbf{w},\mathbf{v}\in\mathbb R^d$ and $a,b\in\mathbb R$, define
\[
k(\mathbf{w},a,\mathbf{v},b)
:=\mathbb E_{\mathbf{x}\sim\mathcal N(\mathbf{0},I_d)}\!\left[
\sigma(\langle\mathbf{w},\mathbf{x}\rangle+a)\,
\sigma(\langle\mathbf{v},\mathbf{x}\rangle+b)
\right].
\]
Assume $\|w\|>0$ and $\|v\|>0$ (the edge cases follow similarly). Set
$\rho=\langle \w,\v\rangle/(\|\w\|\|\v\|)$, $\alpha=-a/\|\w\|$, $\beta=-b/\|\v\|$, and
$s=\sqrt{1-\rho^2}$.  Let $\phi$ and $\Phi$ denote the standard normal density and distribution functions, and let $\Phi_2(\cdot,\cdot;\rho)$ denote the bivariate normal distribution function with correlation $\rho$. Define also the bivariate normal density
\[
\phi_2(u,v;\rho)
=\frac{1}{2\pi\sqrt{1-\rho^2}}
\exp\!\left(-\frac{u^2-2\rho u v+v^2}{2(1-\rho^2)}\right).
\]
Then the kernel admits the explicit expression
\[
\begin{aligned}
	k(\mathbf{w},a,\mathbf{v},b)
	={}&\ \|\mathbf{w}\|\,\|\mathbf{v}\|
	\Bigg(
	\rho\Big[1-\Phi(\alpha)-\Phi(\beta)
	+\Phi_2(\alpha,\beta;\rho)\Big]
	+(1-\rho^2)\,\phi_2(\alpha,\beta;\rho)
	\Bigg)
	\\[0.4em]
	&\quad +\ \|\mathbf{w}\|\,b\;
	\phi(\alpha)\Big(1-\Phi\!\big(\tfrac{\beta-\rho\alpha}{s}\big)\Big)
	+\ \|\mathbf{v}\|\,a\;
	\phi(\beta)\Big(1-\Phi\!\big(\tfrac{\alpha-\rho\beta}{s}\big)\Big)
	\\[0.4em]
	&\quad +\ ab
	\Big[1-\Phi(\alpha)-\Phi(\beta)
	+\Phi_2(\alpha,\beta;\rho)\Big].
\end{aligned}
\]

\begin{remark}[Owen's $T$-function]
The bivariate normal distribution function $\Phi_2$ can also be expressed using Owen's $T$-function, which yields equivalent one-dimensional integral representations that are often employed for numerical evaluation.
\end{remark}
First- and second-order derivatives of the kernel likewise admit explicit expressions. For example, the gradient with respect to $\mathbf{w}$ takes the form
\[
\nabla_{\mathbf{w}}k(\mathbf{w},a,\mathbf{v},b)
= c_1\,\mathbf{w}+c_2\,\mathbf{v},
\]
where
\begin{align*}
\Delta
&\defeq\langle\mathbf{w},\mathbf{w}\rangle\langle\mathbf{v},\mathbf{v}\rangle
-\langle\mathbf{w},\mathbf{v}\rangle^2
=\|\mathbf{w}\|^2\|\mathbf{v}\|^2(1-\rho^2),\\
c_1
&=\frac{\langle\mathbf{v},\mathbf{v}\rangle\,\Gamma_1
	-\langle\mathbf{w},\mathbf{v}\rangle\,\Gamma_2}{\Delta},
\qquad
c_2
=\frac{-\langle\mathbf{w},\mathbf{v}\rangle\,\Gamma_1
	+\langle\mathbf{w},\mathbf{w}\rangle\,\Gamma_2}{\Delta}.\\
\Gamma_1
&=\|\mathbf{w}\|\Bigg(
\|\mathbf{v}\|\Big(
\rho\big[1-\Phi(\alpha)-\Phi(\beta)+\Phi_2(\alpha,\beta;\rho)\big]
+(1-\rho^2)\phi_2(\alpha,\beta;\rho)
\Big)
\\&+b\,\phi(\alpha)\Big(1-\Phi\!\big(\tfrac{\beta-\rho\alpha}{s}\big)\Big)
\Bigg),\\
\Gamma_2
&=\|\mathbf{v}\|\Bigg(
\|\mathbf{v}\|\Big(
1-\Phi(\alpha)-\Phi(\beta)+\Phi_2(\alpha,\beta;\rho)+(1-\rho^2)\,\beta\,\phi(\beta)\Big(1-\Phi\!\big(\tfrac{\alpha-\rho\beta}{s}\big)\Big)
\\&+\rho(1-\rho^2)\phi_2(\alpha,\beta;\rho)
\Big)
+b\,\phi(\beta)\Big(1-\Phi\!\big(\tfrac{\alpha-\rho\beta}{s}\big)\Big)
\Bigg).
\end{align*}

A detailed and systematic analysis of the structural and symmetry properties, together with precise asymptotic characterizations of the eigenspectrum, extending the methodology developed in the present work, and a comparative study of the resulting classes of global and spurious minima associated with the biased kernel, is carried out in \cite{arjevani2026bias}.

\subsubsection{Symmetry breaking and deep architectures}
In general terms, this article developed out of a program to understand how symmetry breaking phenomena exhibited by nonconvex loss landscapes coming from natural distributions may allow gradient-based methods to find good minima efficiently. 
The ability to control the extent to which tangency arcs break the symmetry of critical points as demonstrated in the paper is an instance of a general approach developed in \cite{ArjevaniCP2023} for arguing about tractability of symmetric nonconvex optimization problems.

Critical points of $G$-invariant functions may or may not be symmetric. In \cite{ArjevaniCP2023}, it is shown, however, that when certain conditions apply, critical points connected to symmetric ones by tangency arcs are symmetry breaking. 
Once a lower bound on the isotropy of critical points has been obtained, the effects of increasing the number of neurons on spurious
minima and, crucially, on the possible emergence of descent directions can be
analyzed, e.g., \cite{arjevani2022annihilation}.

A rigorous and systematic analysis of these phenomena in loss landscapes arising from the training of deep neural architectures is developed in \cite{Arjevani2023deepsb}. In that work, the theory of distributions is adopted as a comprehensive analytical framework that permits a precise treatment of the nonsmoothness of ReLU networks. Within this framework, tools from geometric measure theory are employed to obtain a refined characterization of the gradient, interpreted as a vector-valued function of bounded variation, and of the Hessian, which is accordingly represented by Radon measures.

\section{Acknowledgements and disclosure of funding}
We thank Tierra del Sol. The research was supported by the Israel Science Foundation (grant No. 724/22).

\bibliographystyle{abbrv}
\bibliography{bib}

\newpage

\section{The o-minimal structure $\RR_{an}$}\label{sec:omin}

\begin{definition} \label{def:omin}
	An o-minimal structure on $(\RR, +, \cdot)$ is a sequence $\cD = (\cD_n)_{n\in\NN}$ such that for each $n\in\NN$:
	\begin{enumerate}
		\item[(D1)] $\cD_n$ is a boolean algebra of subsets of $\RR^n$, i.e., $\cD_n$ is closed under taking complements and finite unions.
		\item[(D2)] If $A\in\cD_n$, then $A\times \RR$ and $\RR\times A$ are in $\cD_{n+1}$.
		\item[(D3)] If $A\in\cD_{n+1}$ then the projection on the first $n$ coordinates $\pi(A)$ is in $\cD_n$.
		
		\item[(D4)] $\cD_n$ contains $\{\x\in\RR^n~|~P(\x)=0\}$ for every polynomial $P\in \RR[X_1,\dots,X_n]$.
		\item[(D5)] \emph{(o-minimality)} Each set  belonging to $\cD_1$ is a finite union of intervals and points.
\end{enumerate}
\end{definition}
The assertion made in the main text concerning the definability of sets defined by first-order formulae ranging over definable sets should be clearer in light of axioms D1-5. We refer the reader to \cite[Section A]{van1996geometric} for more details. Standard operations such as taking the inverse of a definable function or adding, multiplying and composing definable functions are easily expressed in terms of first-order formulae and so are seen to preserve definability.

By a direct argument, modulo a result by Gabrielov concerning
projections of semianalytic sets \cite{gabrielov1968projections}, the collection of globally subanalytic sets given by inverse images of subanalytic sets under the map $\cV_d(\x)$ forms an o-minimal structure \cite{van1986generalization}. The loss function $\ploss$ can be equivalently given as a sum of terms of the form $\varphi(\w,\v) = \frac{1}{\pi}\|\w\|\|\v\|(\sin(\theta) + (\pi-\theta)\cos(\theta))$, where $\theta(\w,\v) = \cos^{-1}\prn{\frac{\w^\top\v}{\|\w\|\|\v\|}}$ and $\w,\v$ rows of $W$ or $T$. Thus, by the preceding discussion, it suffices to show that the factors of $\varphi$ are $\RR_{an}$-definable for any two vectors: Euclidean norms are algebraic, as is the function $(\w,\v)\mapsto\frac{\w^\top\v}{\|\w\|\|\v\|}$ on its domain, hence both are definable in any o-minimal structure. Moreover, every restriction of an analytic function to a compact subanalytic set is $\RR_{an}$-definable. Therefore $\sin$ and $\cos$ are $\RR_{an}$-definable on $[0,\pi]$, as is $\arccos$ on $[-1,1]$ by its definition as the inverse of $\cos$, concluding the argument.

\section{\pfref{lem:tang_eigsp}}
We need the following result from o-minimal theory.\\

\paragraph{Monotonicity theorem} \cite{van1996geometric}
\textit{Let $f:(a,b)\to\RR$ be a definable function, $-\infty\le a<b\le \infty$. For fixed $p\in\NN$, there are $a_0,\dots,a_{k+1}$ with $a=a_0<a_1<\cdots<a_{k+1} = b$ such that $f|_{(a_i,a_{i+1})}$ is $C^p$, and either constant or strictly monotone for $i=0,\dots,k$.}
\\

\noindent Let $\gamma$ be a definable tangency arc relative to $\c$ parameterized by arc length, thus $\|(\gamma(t) - \c)/t\| = 1,~t>0$. By the monotonicity theorem, $\dot{\gamma}(0)$, as a one-sided limit, exists and is finite. By definition, 
\begin{align}\label{eqn:arc_square}
	\nabla f(\gamma(t)) = \lambda(t) (\gamma(t)-\c),~t>0.
\end{align}
Since $\lambda(t) = (\gamma(t)-\c)^\top \nabla f(\gamma(t)) /\|\gamma(t)-\c\|^2$, $\lambda(t)$ is definable and bounded by the operator norm of $\nabla^2 f(\c)$ and by $\nabla^2 f$ being $C^2$, and so by the monotonicity theorem again $\lambda(0) \defeq \lim_{t\to0^+}\lambda(t)$ exists and is finite. At $t=0$, $\nabla f(\gamma(0))=0$ and so dividing (\ref{eqn:arc_square}) by $t$ and taking a limit, we have 
$\nabla^2 f(\c) \dot{\gamma}(0) = \lambda(0)\dot{\gamma}(0)$, concluding the first part of the lemma.\\

We prove the second part of the lemma for $m(r)$. The $M(r)$ case follows similarly. The existence of a tangency arc $\gamma(t)$ parameterized by arc length   giving $m(t)$ is a consequence of the set of minimizers $\minimizers$ being definable and use of the CSL for $\c\in\cl{\minimizers}$, see \pref{corr:mr_exists}. By the monotonicity theorem, $\gamma$ is $C^1$ ($t>0$ sufficiently small). In addition, parameterization by arc implies  $(\gamma(t)-\c)^\top\dot\gamma(t) = t$. We show that $\dot{\gamma}(0)$ is an eigenvector associated to the minimal eigenvalue. As $f$ is $C^2$, we may write
\begin{align*}
f(\x) &= f(\c) + \nabla f(\c)^\top(\x-\c) + (\x-\c)^\top\nabla^2 f(\c)(\x-\c) + 
(\x-\c)^\top H(\x)(\x-\c)\\ &= 
f(\c) + (\x-\c)^\top(\nabla^2 f(\c) + H(\x))(\x-\c),
\end{align*}
with $H$ symmetric matrix-valued function such that $\lim_{\x\to\c} H(\x) = 0$, and similarly
\begin{align*}
f\circ \gamma(t) &= f\circ \gamma(0) + (f\circ \gamma)'(t)t + h(t)t^2\\
&= f(\c) + \nabla f(\gamma(t))\dot\gamma(t)t + h(t)t^2\\
&= f(\c) + \lambda(t)(\gamma(t)-\c)\dot\gamma(t)t + h(t)t^2\\
&= f(\c) + (\lambda(t) + h(t))t^2,
\end{align*}
with $\lambda$ as in (\ref{eqn:arc_square}) and $h$ a real-valued function such that $\lim_{t\to0} h(t) = 0$. Combined, and letting $\eta_1(A)$ denote the minimal eigenvalue of a symmetric matrix $A$, we obtain
\begin{align*}
f(\c) + (\lambda(t) + h_2(t))t^2 
&= 	\min_{\|\x-\c\|=t} \crl{f(\c) + (\x-\c)^\top(\nabla^2 f(\c) + H(\x))(\x-\c)}\\
&= f(\c) + \eta_1[(\nabla^2 f(\c) + H(\x_m(t)))]t^2,
\end{align*}
where $\x_m(r)$ is any point in $X^m_\c$ at distance $t$ from $\c$. By standard results from eigenvalue perturbation theory (e.g., Weil's inequality), $\eta_1[(\nabla^2 f(\c) + H(\x_m)I_d)]\to \eta_1[(\nabla^2 f(\c)]$, thus $\lambda(0)= \eta_1[(\nabla^2 f(\c)]$. Proceeding as in the proof of the first part of the lemma shows that $\dot\gamma(0)$ is an eigenvector associated with  $\eta_1[(\nabla^2 f(\c)]$.

\section{\pfref{thm:max_iso}} \label{sec:thm_max_iso_prf}

We begin with providing an explicit description of the isotypic components of $(M(d,d), S_d)$ following the parameterization given in \cite{arjevanifield2020hessian,arjevanifield2021analytic,arjevani2022annihilation}.

Let $d > 1$. Take the natural (orthogonal) action of $S_d$ on $\real^d$ defined by permuting coordinates.  The representation is not irreducible since the subspace $E = 
\{(x,x,\cdots,x)\in\real^d \dd x \in \real\}$ 
is 	invariant by the action of $S_d$, as is the hyperplane 
$$H_{n-1} = E^\perp =\{(x_1,\cdots,x_d)\dd \sum_{i\in\ibr{n}}x_i =0\}.$$
It is easy to check that $(E,S_d)$, also called the \emph{trivial }representation of $S_d$, and $(H_{d-1},S_d)$, the \emph{standard} representation, are irreducible, real, and 
not isomorphic.	Their isomorphism classes are denoted by $\mathfrak{t}$ and $\mathfrak{s}_{d}$, respectively (for all $d \ge 2$, $\mathfrak{t}$ is 1-dimensional). 
\begin{lemma}[\cite{arjevanifield2020hessian}]\label{lem: isotD}
Write $	M(d, d) = \bD_d \oplus \bS_d \oplus \bA_d$, $\bD_d, \bS_d, \bA_d$. Then,
for $d\ge 4$,
\begin{itemize}[leftmargin=*]
\item 		
$\mathbb{D}_{d}$ is the orthogonal $S_d$-invariant direct sum $\mathbb{D}_{d,1} \oplus  \mathbb{D}_{d,2}$,
where 
\begin{enumerate}[leftmargin=*]
	\item $\mathbb{D}_{d,1} $ is the space of diagonal matrices with all entries equal and is naturally isomorphic to $(E,S_d)$. 
	\item $\mathbb{D}_{d,2}$ is the $(d-1)$-dimensional space of diagonal matrices with diagonal entries summing to zero and is naturally isomorphic to $(H_{d-1},S_d)$.
\end{enumerate}
In particular, the isotypic decomposition of $(\mathbb{D}_d,S_d)$ is $\mathfrak{t}+\mathfrak{s}_d$.

\item $\mathbb{A}_{d}$ is the orthogonal $S_d$-invariant direct sum $\mathbb{A}_{d,1} \oplus  \mathbb{A}_{d,2}$, where
\begin{enumerate}
	\item $\mathbb{A}_{d,1}$ is the $(d-1)$-dimensional space of matrices $[a_{ij}]$ for which there exists $(x_1,\cdots,x_d) \in H_{d-1}$ such that
	for all $i,j \in \ibr{d}$,  $a_{ij} = x_i - x_j$, 
	\item  $\mathbb{A}_{d,2}$ consists of all skew-symmetric matrices with row sums zero.
\end{enumerate}
As representations, $(\mathbb{A}_{d,1},S_d)$ is isomorphic to $(H_{k-1},S_d)$ and $(\mathbb{A}_{d,2},S_d)$ is isomorphic to $(\wedge^2 H_{d-1},S_d)$.
In particular, the isotypic decomposition of $(\mathbb{A}_d,S_d)$ is $\mathfrak{s}_d + \mathfrak{x}_d$.  

\item $\mathbb{S}_d$ is the orthogonal $S_d$-invariant direct sum $\mathbb{S}_{1,d} \oplus  \mathbb{S}_{2,d}\oplus \mathbb{S}_{3,d}$, where
\begin{enumerate}
	\item $\mathbb{S}_{d, 1}$ is the $1$-dimensional space of symmetric matrices with diagonal entries zero and all off diagonal entries equal.
	\item $\mathbb{S}_{d, 2}$ is the $(d-1)$-dimensional space of matrices $[a_{ij}]\in \mathbb{S}_d$ for which there exists $(x_1,\cdots,x_d) \in H_{d-1}$ such that
	for all $i,j \in \ibr{d}$, $i \ne j$,  $a_{ij} = x_i + x_j$.
	\item $\mathbb{S}_{d, 3}$ consists of all symmetric matrices in $\mathbb{S}_d$ with all row (equivalently, column) sums zero.
	\item $\text{dim}(\mathbb{S}_{d, 3}) = \frac{d(d-3)}{2}$.
\end{enumerate}
The representations $(\mathbb{S}_{d,i},S_d)$ are irreducible, $i \in \ibr{3}$:  $(\mathbb{S}_{d,1},S_d)$ is isomorphic to the trivial representation,
$(\mathbb{S}_{d,2})$ is isomorphic to the standard representation and
$(\mathbb{S}_{d,3},S_d)$ is isomorphic to the $S_d$-representation associated to the partition $(d-2,2)$ (isomorphism type $\mathfrak{y}_d$).
\end{itemize}
\end{lemma}
We collect the sub-representations constituting $\bD_d, \bS_d$ and $\bA_d$ by their isomorphism type and prove \pref{thm:max_iso} case by case.
\\
\paragraph{The isotypic component $\bV_\ft$.}
We have $\bV_\ft= \bD_{d,1} + \bS_{d,1} = M(d,d)^{\Delta S_d}$, concluding the trivial part of the theorem.
\\
\paragraph{The isotypic component $\bV_\fs$.}
The three sub-representations in $(M(d,d),S_d)$ identified as isomorphic to the standard are 
\begin{equation} \label{eqn:std_dec}
\bV_\fs = \bD_{d,2} + \bS_{d,2} + \bA_{d,1}.
\end{equation}
Computing, we see that $\bV_\fs \cap M(d,d)^{\Delta S_d} = 0$ and
$\dim\prn{\bV_\fs \cap M(d,d)^{\Delta (S_{d-1}\times S_1)}} = 3$ with 
\begin{align}\label{std_struc}
\bD_{d,2} \cap M(d,d)^{\Delta S_{d-1}} &= \frac{-\alpha}{d-1}I_{d-1}\oplus [\alpha],\nonumber\\
\bS_{d,2} \cap M(d,d)^{\Delta S_{d-1}} &= 
\begin{bmatrix}
\frac{-2}{d-2}\alpha(\bones_{d-1} \bones_{d-1}^\top - I_{d-1})& \alpha\bones_{d-1},\\
\alpha\bones_{d-1}^\top& 0
\end{bmatrix},
\\
\bA_{d,1} \cap M(d,d)^{\Delta S_{d-1}} &= 
\begin{bmatrix}
0_d& -\alpha\bones_{d-1}\\
\alpha\bones_{d-1}^\top& 0
\end{bmatrix}.\nonumber
\end{align}
This proves that the maximal isotropy type in $\bV_\fs$ is $(S_{d-1}\times S_1)$. 

Let $\rho_i$ denote the $S_d$-isomorphism from $H_{d-1}$ to the $i$th factor in (\ref{eqn:std_dec}) giving the explicit parameterizations described in \pref{lem: isotD} (namely, $\rho_1:H_{d-1}\to \bD_{d,2}$, $\rho_2:H_{d-1}\to \bS_{d,2}$ and $\rho_3:H_{d-1}\to \bA_{d,1}$). Thus for example, if  $(x_1,\dots,x_d)\in H_{d-1}$ then
\begin{equation}
\rho_1(x_1,\dots,x_d) = 
\begin{bmatrix}
x_1\\&&\ddots\\&&&x_d
\end{bmatrix}.
\end{equation}
Let $\pi_i$ denote the projection of $\bV_\fs$ on the $i$th factor
in (\ref{eqn:std_dec}), and set $A_{ij} \defeq\rho_j^{-1}\pi_j A\rho_i: H_{d-1}\to H_{d-1},~i,j \in [3]$. Being a composition of $S_d$-maps, $A_{ij}$ is an $S_d$-self map on $H_{d-1}$ and so is simply a multiplication by some scalar $\alpha_{ij}$. Set $\x= (1,1,\dots,1,-(d-1))\in H_{d-1}$ and let $\fS_i \defeq \rho_i(x)$. We have $A(\fS_i) = \sum_{j} \pi_j A(\fS_i) = \sum_{j} \pi_j A \rho_i(x) = \sum_{j} \rho_j A_{ij}(x) = \sum_{j} \rho_j (\alpha_{ij}x) = \sum_{j} \alpha_{ij}\fS_j$.  Therefore, the three eigenvalues of $[\alpha_{ij}]$ give the eigenvalues of $A|_{\bV_\fs}$ with multiplicities multiplied by $d-1$. Being an $S_d$-map, $A$ is clearly an $(S_{d-1}\times S_1)$-map and so restricts to a self-map on $M(d,d)^{\Delta (S_{d-1}\times S_1)}$. The vector $\x$, intentionally chosen so that $\rho_i(\x)\in M(d,d)^{S_{d-1}}\cap \bV_\fs$ for all $i\in[3]$, implies in a similar vein that the eigenvalues of $[\alpha_{ij}]$ are the eigenvalues of $A|_{\bV_\fs\cap M(d,d)^{\Delta (S_{d-1}\times S_1)}}$, this time with the same multiplicity, concluding the $\fs$-case. 
\\
\paragraph{The isotypic component $\bV_\fx$.}
The single sub-representation in $(M(d, d),S_d)$ identified as isomorphic to the exterior square is
\begin{equation} \label{eqn:x_dec}
	\bV_\fx =  \bA_{d,2}
\end{equation}
In particular, there is a single $\fx$-eigenvalue. Its multiplicity is $(d-1)(d-2)/2$. Computing, we see that $\bV_\fx \cap M(d,d)^{\Delta S_d} = \bV_\fx \cap M(d,d)^{\Delta (S_{d-1}\times S_1)} = 0$ and
$\dim\prn{\bV_\fx \cap M(d,d)^{\Delta (S_{d-2}\times S_1^2)}} = 1$ with $\bV_\fx \cap M(d,d)^{\Delta (S_{d-2}\times S_1^2)}$ given in the form
\begin{align}\label{eqn:x_struc}
\begin{bmatrix} 
	0_{d-2}& \frac{-\alpha}{d-2}\bones_{d-2}& & \frac{\alpha}{d-2}\bones_{d-2}\\
	\frac{\alpha}{d-2}\bones_{d-2}^\top& 0&-\alpha\\
	\frac{-\alpha}{d-2}\bones_{d-2}^\top& \alpha&0
\end{bmatrix}.
\end{align}
By the same reasoning used in the $\fs$-case, $A|_{\bV_\fx\cap M(d,d)^{\Delta (S_{d-2}\times S_1^2)}}$ has a single eigenvalue, multiplicity one.
\\
\paragraph{The isotypic component $\bV_\fy$.}
The single sub-representation in $(M(d, d),S_d)$ identified as isomorphic to the representation associated to the partition $(d-2,2)$ (referred to here as the $\fy$-representation) is
\begin{equation} \label{eqn:y_dec}
	\bV_\fy =  \bS_{d,3}.
\end{equation}
In particular, there is a single $\fy$-eigenvalue. Its multiplicity is $(d-1)(d-2)/2$. Computing, we see that $\bV_\fy \cap M(d,d)^{\Delta S_d} = \bV_\fy \cap M(d,d)^{\Delta (S_{d-1}\times S_1)} = 0$ and
$\dim\prn{\bV_\fy \cap M(d,d)^{\Delta (S_{d-2}\times S_1^2)}} = 1$ with $\bV_\fy \cap M(d,d)^{\Delta (S_{d-2}\times S_1^2)}$ given in the form
\begin{align*}
	\begin{bmatrix}
		\frac{2\alpha}{(d-2)(d-3)}(\bones_{d-2}\bones_{d-2}^\top - I_{d-2})& \frac{-\alpha}{d-2}\bones_{d-2}& & \frac{-\alpha}{d-2}\bones_{d-2}\\
		\frac{-\alpha}{d-2}\bones_{d-2}^\top& 0&\alpha\\
		\frac{-\alpha}{d-2}\bones_{d-2}^\top& \alpha&0
	\end{bmatrix}.
\end{align*}
By the same reasoning used for the $\fs$-case, $A|_{\bV_\fy\cap M(d,d)^{\Delta (S_{d-2}\times S_1^2)}}$ has a single eigenvalue, multiplicity one.

\section{\pfref{thm:type_iso}}

As mentioned in the introductory part, the Hessian spectrum of type I and  II critical points agree modulo $O(d^{-1/2})$-terms
\cite{arjevanifield2020hessian,arjevanifield2021analytic,arjevani2022annihilation}. To identify the isotypic components giving the minimal eigenvalue we compute two leading terms of the Puiseux series describing the Hessian eigenvalues. This requires the development of Puiseux series describing the entries of type I and II minima to higher order terms.
The results given in \pref{corr:eig_sym} adapt to $\Delta (S_{d-p}\times S_p)$-maps as follows. Assume $p+q = d$, regard $S_p \times S_q$ as a subgroup of $S_d$ and restrict the diagonal action of $S_d$ on $M(d,d)$ to $S_p \times S_q$ to define $M(d,d)$ as an $S_p \times S_q$-space. We assume $d >p > d/2$ so that $S_p \times S_q$ will be a maximal intransitive subgroup of $S_d$~\cite{arjevanifield2019spurious}. Clearly, $M(d,d)$ decomposes as an orthogonal $S_p \times S_q$-invariant direct sum
\[
M(d,d) = M(p,p) \oplus M(p,q) \oplus M(q,p) \oplus M(q,q),  
\]
where $M(p,p)$ is an $S_p$-space and $M(q,q)$ is an $S_q$ space (diagonal actions). We regard $M(p,q)$  and $M(q,p)$ as $S_p \times S_q$-spaces. Thus, $S_p$ acts on $M(p,q)$ (resp.~$M(q,p)$) by permuting rows (resp.~columns) and $S_q$ acts on $M(p,q)$ (resp.~$M(q,p)$) by permuting columns (resp.~rows).

Initial terms of Puiseux series of type I and type II minima were given in previous work \cite{ArjevaniField2020,arjevanifield2021analytic,arjevani2022annihilation}. However, as mentioned in the main text, higher order terms are required for the computation of $O(d^{-1/2})$-eigenvalue terms. 
By the orthogonal symmetry of the Gaussian distribution, results hold for any $T$ determined by a matrix in $O(\RR^d)$.
For simplicity, assume $T=I$.  Further simplification is given by a straightforward reduction used in \cite{arjevanifield2021analytic}: the object of study being the loss landscape, rather than optimization processes, we may set as ones the second layer of weights, $\ba$ and $\bb$, for the critical points considered without loss of generality.

\end{document}